\documentclass[runningheads]{llncs}
\usepackage{graphicx}
\usepackage{caption}
\usepackage{floatrow}
\usepackage{subcaption}
\usepackage{booktabs}
\usepackage{adjustbox}
\usepackage[symbol]{footmisc}

\usepackage{tikz}
\usepackage{comment}
\usepackage{amsmath,amssymb} 
\usepackage{color}
\usepackage{booktabs}
\usepackage{multirow}
\usepackage{cite}
\usepackage{tablefootnote}

\usepackage{amssymb}
\usepackage{pifont}
\newcommand{\xmark}{\ding{53}}%

\usepackage[accsupp]{axessibility}  

\providecommand{\anish}[1]{{\protect\color{purple}{[Anish: #1]}}}

\begin{document}
\pagestyle{headings}
\mainmatter

\def\ECCVSubNumber{5750}  

\title{SPIN: An Empirical Evaluation on Sharing Parameters of Isotropic Networks}

\titlerunning{SPIN}

\author{Chien-Yu Lin\inst{1}\thanks{Equal contribution.}\thanks{Work done while interning at Apple.} \and
Anish Prabhu\inst{2}$^{*}$ \and
Thomas Merth\inst{2} \and
Sachin Mehta\inst{2} \and
Anurag Ranjan\inst{2} \and
Maxwell Horton\inst{2} \and
Mohammad Rastegari\inst{2}}

\authorrunning{C. Lin et al.}
%

\institute{
University of Washington, USA \\
\and
Apple, Inc., USA\\
}

\maketitle

\begin{abstract}
Recent isotropic networks, such as ConvMixer and Vision Transformers, have found significant success across visual recognition tasks, matching or outperforming non-isotropic Convolutional Neural Networks. Isotropic architectures are particularly well-suited to cross-layer weight sharing, an effective neural network compression technique. In this paper, we perform an empirical evaluation on methods for sharing parameters in isotropic networks (SPIN). We present a framework to formalize major weight sharing design decisions and perform a comprehensive empirical evaluation of this design space. Guided by our experimental results, we propose a weight sharing strategy to generate a family of models with better overall efficiency, in terms of FLOPs and parameters versus accuracy, compared to traditional scaling methods alone, for example compressing ConvMixer by $1.9\times$ while improving accuracy on ImageNet. Finally, we perform a qualitative study to further understand the behavior of weight sharing in isotropic architectures. The code is available at \url{https://github.com/apple/ml-spin}.

\end{abstract}

\section{Introduction}
\begin{figure}[th!]
    \centering
    \begin{subfigure}{\linewidth}
        \includegraphics[width=\linewidth]{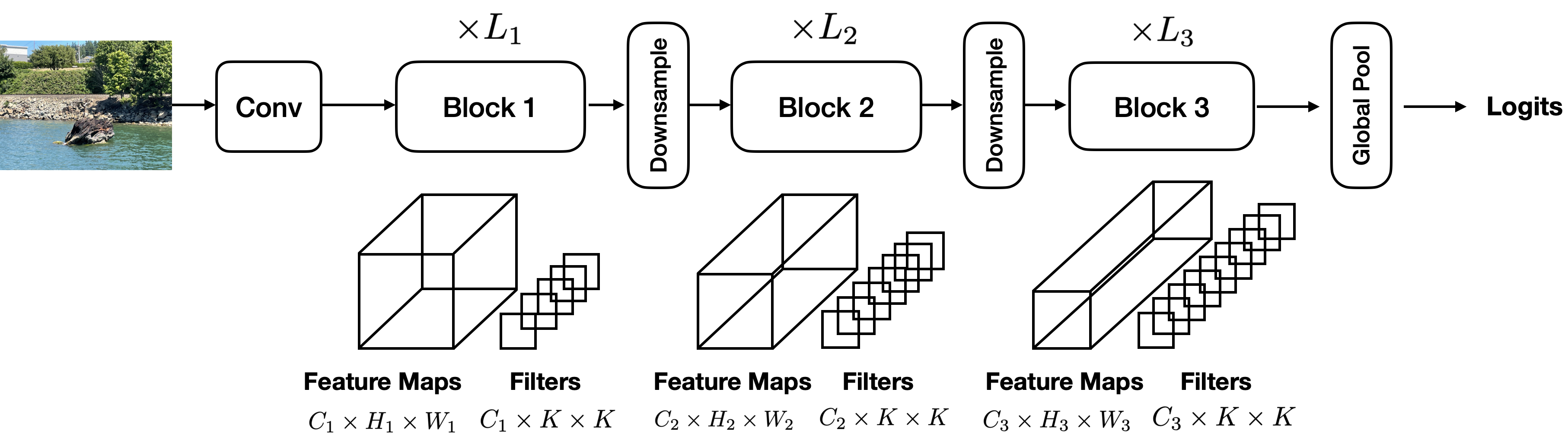}
        \caption{Architecture of regular CNNs.}
        \label{subfig:cnn_arch}
    \end{subfigure}
    \begin{subfigure}{0.65\linewidth}
        \includegraphics[width=\linewidth]{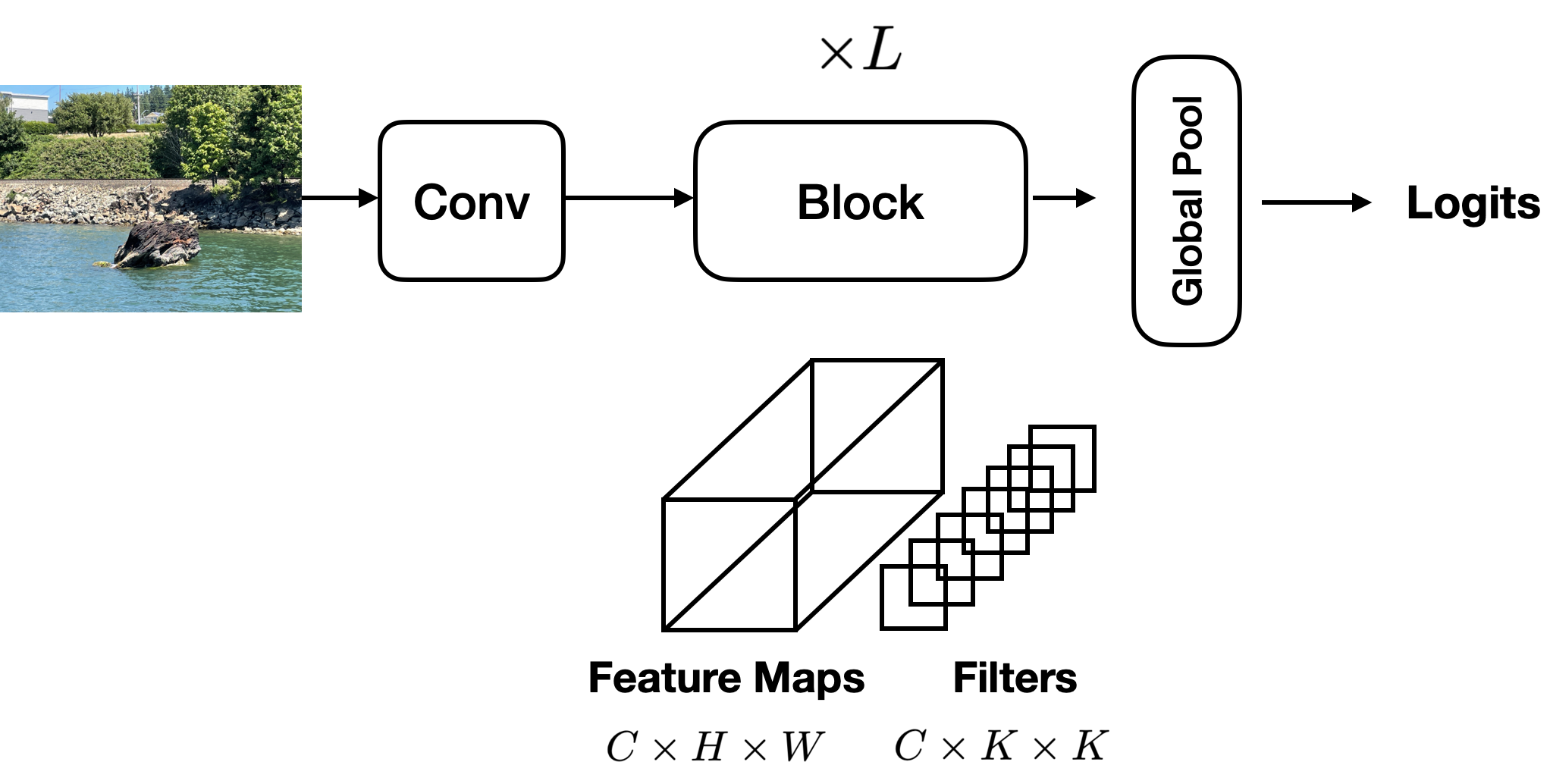}
        \caption{Architecture of isotropic CNNs.}
        \label{subfig:iso_cnn_arch}
    \end{subfigure}
    \caption{Basic architectures of regular and isotropic CNNs. (a) Regular CNNs vary the shape of intermediate features and weight tensors in the network while (b) isotropic CNNs fix the shape of all intermediate features and weight tensors in the network.}
    \label{fig:arch_compare}
\end{figure}

Isotropic neural networks have the property that all of the weights and intermediate features have identical dimensionality, respectively (see Figure \ref{fig:arch_compare}). Some notable convolutional neural networks (CNNs) with isotropic structure \cite{convmixer,convnext} have been proposed recently in the computer vision domain, and have been applied to different visual recognition tasks, including image classification, object detection, and action recognition. These isotropic CNNs contrast with the typical ``hierarchical'' design paradigm, in which spatial resolution and channel depth are varied throughout the network (e.g., VGG \cite{vgg} and ResNet \cite{resnet}).

The Vision Transformer (ViT) \cite{dosovitskiy2020image} architecture also exhibits this isotropic property, although softmax self-attention and linear projections are used for feature extraction instead of spatial convolutions. Follow-up works have experimented with various modifications to ViT models (e.g. replacing softmax self-attention with linear projections \cite{mlp-mixer}, factorized attention \cite{aft}, and  non-learned transformations \cite{vitshift}); however, the isotropic nature of the network is usually retained.

Recent isotropic models (e.g., ViT \cite{dosovitskiy2020image}, ConvMixer \cite{convmixer}, and ConvNext \cite{convnext}) attain state-of-the-art performance for visual recognition tasks, but are computationally expensive to deploy in resource constrained inference scenarios. In some cases, the parameter footprint of these models can introduce memory transfer bottlenecks in hardware that is not well equipped to handle large amounts of data (e.g. microcontrollers, FPGAs, and mobile phones) \cite{albert}. Furthermore, ``over-the-air'' updates of these large models can become impractical for continuous deployment scenarios with limited internet bandwidth. Parameter (or weight) sharing\footnote{We interchangeably use the terms parameter and weight sharing throughout this paper.}, is one approach which compresses neural networks, potentially enabling the deployment of large models in these constrained environments.

Isotropic DNNs, as shown Figure \ref{fig:arch_compare}, are constructed such that a layer's weight tensor has identical dimensionality to that of other layers. Thus, cross-layer parameter sharing becomes a straightforward technique to apply, as shown in ALBERT \cite{albert}. On the other hand, weight tensors within non-isotropic networks cannot be shared in this straightforward fashion without intermediate weight transformations (to coerce the weights to the appropriate dimensionality). In Appendix A, we show that the search space of possible topologies for straightforward cross-layer parameter sharing is significantly larger for isotropic networks, compared to ``multi-staged'' networks (an abstraction of traditional, non-isotropic networks). This rich search space requires a comprehensive exploration. Therefore, in this paper, we focus on isotropic networks, with the goal of finding practical parameter sharing techniques that enable high-performing, low-parameter neural networks for visual understanding tasks. 

To extensively explore the weight sharing design space for isotropic networks, we experiment with different orthogonal design choices (Section \ref{subsec:design-space-exploration}). Specifically, we explore (1) different sharing topologies, (2) dynamic transformations, and (3) weight fusion initialization strategies from pretrained non-sharing networks. Our results show that parameter sharing is a simple and effective method for compressing large neural networks versus standard architectural scaling approaches (e.g. reduction of input image size, channel size, and model depth). Using a weight sharing strategy discovered from our design space exploration, we achieve nearly identical accuracy (to non-parameter sharing, iso-FLOP baselines) with significantly reduced parameter counts. Beyond the empirical accuracy versus efficiency experiments, we also investigate network representation analysis (Section \ref{sec:representation_analysis}) and model generalization (Appendix F) for parameter sharing isotropic models.
\setcounter{secnumdepth}{1}
\section{Related Works}

\begin{figure}[th!]
    \centering
    \begin{subfigure}{\linewidth}
        \includegraphics[width=\linewidth]{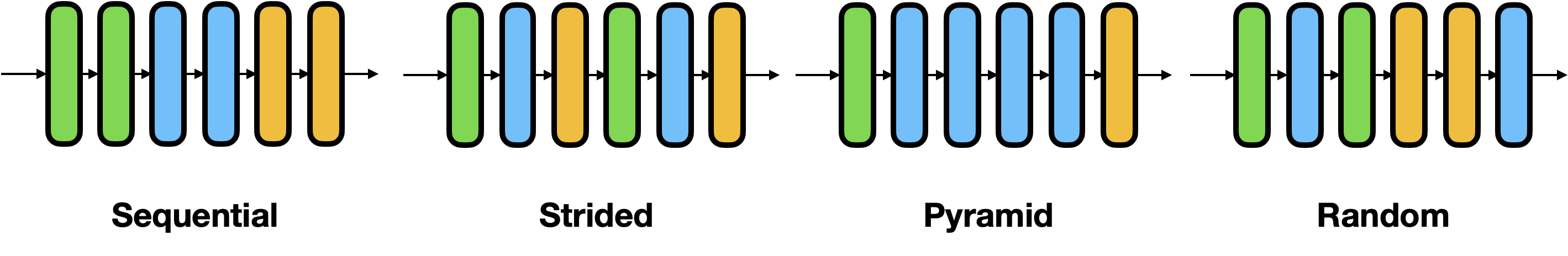}
        \caption{Sharing mapping.}
        \label{subfig:sharing_topology_map}
    \end{subfigure}
    \begin{subfigure}{\linewidth}
        \includegraphics[width=\linewidth]{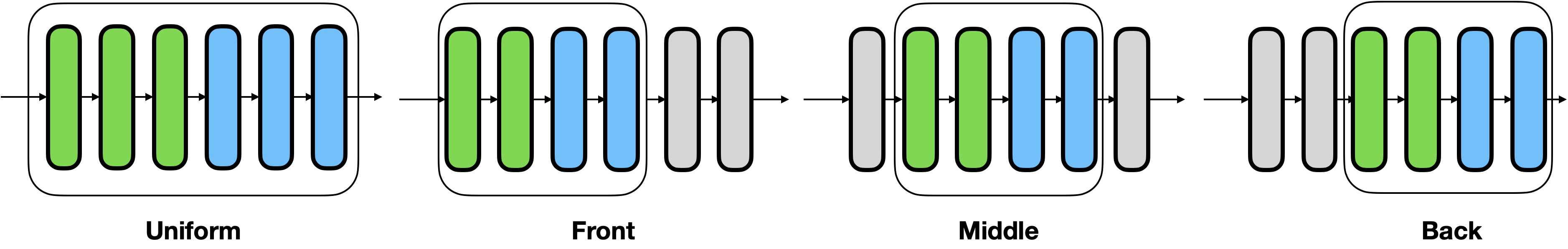}
        \caption{Sharing distribution.}
        \label{subfig:sharing_topology_dist}
    \end{subfigure}
    \caption{\textbf{Sharing topologies}. In (a), sharing mapping determines which layers share the same weights while in (b), sharing distribution determines how the weight sharing layers are distributed in the network.  Layers with the same color share weights. Layers outside of the sharing section do not share weights. Best viewed in color.  
    }
    \label{fig:sharing_topology}
\end{figure}

\paragraph{\bfseries Cross-layer Parameter Sharing.} Cross-layer parameter sharing has been explored for both CNN- and Transformer-based models \cite{Kubilius2018AligningAN,brainrnn,Spoerer677237,deeplyshared,Dehghani2019UniversalT,albert,cyclerev}. For instance, Kim et al. \cite{deeplyshared} applies cross-layer parameter sharing across an entire heterogeneous CNN. However, they share weights at the granularity of filters, whereas we share weights at the granularity of layers. In terms of our framework, Kubilius et al. \cite{Kubilius2018AligningAN} experiments with Uniform-Strided, proposing a heterogeneous network based off of the human visual cortex. With isotropic networks, we can decouple parameter sharing methods from the constraints imposed by heterogeneous networks. Thus, we expand the scope of weight sharing structures from their work to isotropic networks.

Cross-layer parameter sharing is explored for isotropic Transformer models for the task of neural language modeling \cite{Dehghani2019UniversalT,albert} and vision \cite{cyclerev}. Lan et al. \cite{albert} experiments with Uniform-Sequential, and Dehghain et al. \cite{Dehghani2019UniversalT} experiments with universal sharing (i.e. all layers are shared). Takase et al. \cite{cyclerev} experiments with 3 strategies, namely Uniform-Sequential, Uniform-Strided, and Cycle. In this paper, we extend these works by decomposing the sharing topology into combinations of different sharing mappings (Figure \ref{subfig:sharing_topology_map}) and sharing distributions (Figure \ref{subfig:sharing_topology_dist}).

\paragraph{\bfseries Dynamic Recurrence for Sharing Parameters.} Several works \cite{adaptivelateral,iamnn,Guo2019DynamicRN,slicedrecursive} explore parameter sharing through the lens of dynamically repeating layers. However, each technique is applied to a different model architecture, and evaluated in different ways. Thus, without a common framework, it's difficult to get a comprehensive understanding of how these techniques compare. While this work focuses only on static weight sharing, we outline a framework that may encompass even these dynamic sharing schemes. In general, we view this work as complementary to explorations on dynamic parameter sharing, since our analysis and results could be used to help design new dynamic sharing schemes.

\setcounter{secnumdepth}{2}

\setcounter{secnumdepth}{2}

\section{Sharing Parameters in Isotropic Networks}
\label{parameter_sharing_methods}
In this section, we first motivate why we focus on isotropic networks for weight sharing (Section \ref{subsec:why_iso}), followed by a comprehensive design space exploration of methods for weight sharing, including empirical results (Section \ref{subsec:design-space-exploration}).

\subsection{Why Isotropic Networks?}
\label{subsec:why_iso}
Isotropic networks, shown in Figure \ref{subfig:iso_cnn_arch}, are simple by design, easy to analyze, and enable flexible weight sharing, as compared to heterogeneous networks.

\paragraph{\bfseries Simplicity of Design.} Standard CNN architectural design, whether manual \cite{resnet, Sandler2018MobileNetV2IR}) or automated through methods like neural architecture search \cite{Tan2019MnasNetPN, Howard2019SearchingFM}), require searching a complex search space, including what blocks to use, where and when to downsample the input, and how the number of channels should vary throughout the architecture. On the other hand, isotropic architectures form a much simpler design space, where just a single block (e.g., attention block in Vision Transformers or convolutional block in ConvMixer) along with network's depth and width must be chosen. The simplicity of implementation for these architectures enables us to more easily design generic weight sharing methods across various isotropic architectures. The architecture search space of these networks is also relatively smaller than non-isotropic networks, which makes them a convenient choice for large scale empirical studies. 

\paragraph{\bfseries Increased Weight Sharing Flexibility.} Isotropic architectures provide significantly more flexibility for designing a weight sharing strategy than traditional networks.

We define the \textit{sharing topology} to be the underlying structure of how weight tensors are shared throughout the network. Suppose we have an isotropic network with $L \geq 1$ layers and a weight tensor ``budget'' of $ 1 \leq P \leq L$. The problem of determining the optimal sharing topology can be seen as a variant of the set cover problem; we seek a set cover with no more than $B$ disjoint subsets, which maximizes the accuracy of the resulting network. More formally, a possible sharing topology is an ordered collection of disjoint subsets $\mathcal{T} = (\mathcal{S}_1, \mathcal{S}_2, ..., \mathcal{S}_P)$, where $\cup_{i=1}^B \mathcal{S}_i = \{1, 2, ..., L\}$ for some $1 \leq P \leq L$. We define $\frac{L}{P}$ to be the \emph{share rate}.

We characterize the search space in Appendix A, showing that isotropic networks support significantly more weight sharing topologies than heterogeneous networks (when sharing at the granularity of weight tensors). This substantially increased search space may yield more effective weight sharing strategies in isotropic networks than non-isotropic DNNs, a reason why we are particularly interested in isotropic networks

\paragraph{\bfseries Cross-layer Representation Analysis.} 
To better understand if the weights of isotropic architectures are amenable to compression through weight sharing, we study the representation of these networks across layers. We hypothesize that layers with similar output representations will be more compressible via weight sharing. To build intuition, we use Centered Kernel Alignment (CKA) \cite{Kornblith2019SimilarityON}, a method that allows us to effectively measure similarity across layers. 

\begin{figure}[th!]
    \centering
    \includegraphics[width=1.0\linewidth]{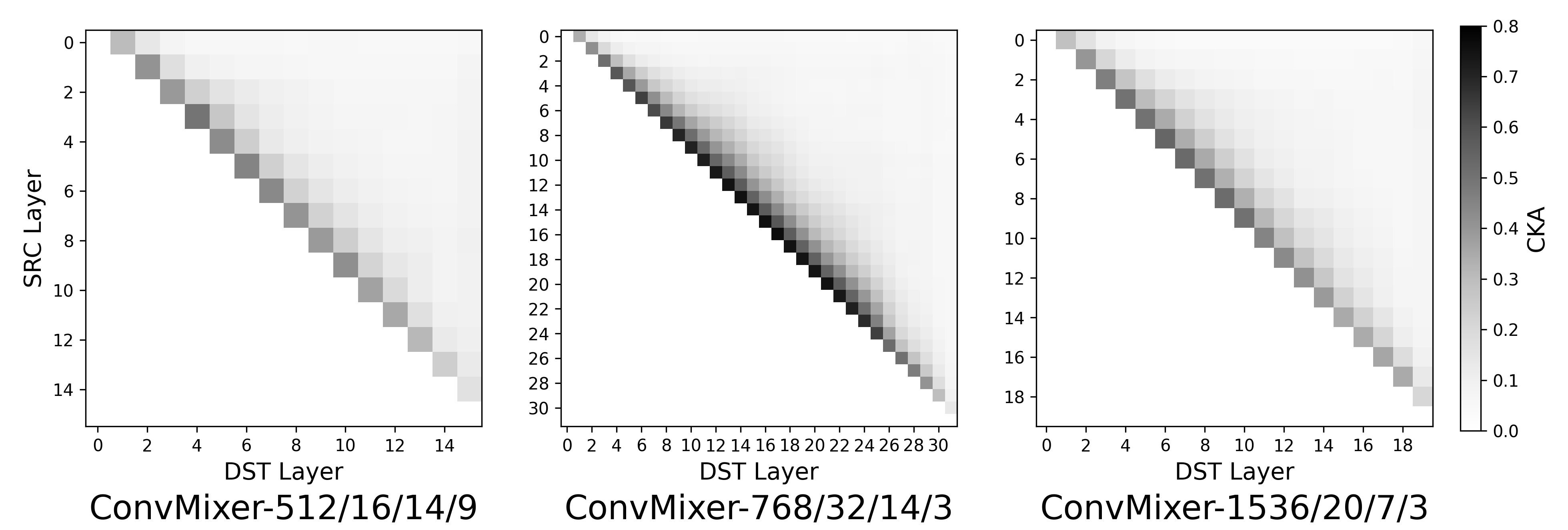}
    \caption{CKA similarity analysis on ConvMixer's intermediate feature maps shows that the output feature maps of neighboring layers and especially the middle layers have the highest similarity. Here, we compute the CKA similarity of each layer's output feature maps. The diagonal line and the lower triangle part are masked out for clarity. The CKA for the diagonal line is 1 since they are identical. The CKA for the lower triangle is the mirror of the upper triangle. Best viewed on screen.}
    \label{fig:convmixer_fmap_cka}
\end{figure}

Figure \ref{fig:convmixer_fmap_cka} shows the pairwise analysis of CKA across layers within the ConvMixer network. We find significant representational similarity for nearby layers. This is not unexpected, given the analysis of prior works on iterative refinement in residual networks \cite{iterative}. Interestingly, we find that CKA generally peaks in the middle of the network for different configurations of ConvMixer. Overall, these findings suggest that isotropic architectures may be amenable to weight sharing, and we use this analysis to guide our experiments exploring various sharing topologies in Section \ref{subsec:design-space-exploration}.

\subsection{Weight Sharing Design Space Exploration} 
\label{subsec:design-space-exploration}
When considering approaches to sharing weights within a neural network, there is an expansive design space to consider. This section provides insights as well as empirical evaluation to help navigate this design space. We first consider the weight sharing topology. Then, we introduce lightweight dynamic transformations on the weights to increase the representational power of the weight-shared networks. Finally, we explore how to use the trained weights of an uncompressed network to further improve accuracy in weight-sharing isotropic networks. All experiments done in this section are based on a ConvMixer model with 768 channels, depth of 32, patch extraction kernel size of 14, and convolutional kernel size of 3.

\paragraph{\bfseries Weight Sharing Topologies.} Isotropic networks provide a vast design space for sharing topologies. We perform an empirical study of various sharing topologies for the ConvMixer architecture, evaluated on the ImageNet dataset. We characterize these topologies by the (1) \textit{sharing mapping} (shown in Figure \ref{subfig:sharing_topology_map}), which describes the structure of shared layers, and (2) the \textit{sharing distribution} (shown in Figure \ref{subfig:sharing_topology_dist}), which describes which subset of layers sharing is applied to. We study the following sharing mappings:

\begin{enumerate}
  \item \textbf{Sequential}: Neighboring layers are shared in this topology. There is motivated by our cross layer similarity analysis in Section \ref{subsec:why_iso} and Figure \ref{fig:convmixer_fmap_cka}, which suggest that local structures of recurrence may be promising.
  \item \textbf{Strided}: This topology defines the recurrence on the network level rather than locally. If we consider having $P$ blocks with unique weights, we first run all of the layers sequentially, then we repeat this whole structure $L / P$ times.
  \item \textbf{Pyramid}: This topology is an extension of Sequential, which has increasingly more shared sequential layers as you approach the center of the network. This is inspired by (1) empirical results in Figure \ref{fig:convmixer_fmap_cka} that show a similar structure in the layer-wise similarity and (2) neural network compression methods (e.g. quantization and sparsity methods), which leave the beginning and end of the network uncompressed \cite{Han2015LearningBW,Rastegari2016XNORNetIC}.
  \item \textbf{Random}: We randomly select which layers are shared within the network, allowing us to understand how much the choice of topology actually matters.
\end{enumerate}

For the sharing distribution, we consider applying (1) \textbf{Uniform}, where sharing mapping is applied to all layers, (2) \textbf{Front}, where sharing mapping is applied to the front of the network, (3) \textbf{Middle}, where sharing mapping is applied to the middle of the network, and (4) \textbf{Back}, where sharing mapping is applied to the back of the network. Note that front, middle and back sharing distributions results in a non-uniform distribution of share rates across layers.

\setlength{\tabcolsep}{4pt}
\begin{table}[t!]
    \centering
    \caption{\textbf{Effect of different sharing distributions and mappings on the performance of weight-shared (WS) ConvMixer with a share rate of 2.} In order to maintain the fixed share rate 2 for non-uniform sharing distributions (i.e., Middle, Front and Back), we apply sharing to 8 layers with share rate $3\times$ and have 16 independent layers. For Middle-Pyramid, the network is defined as $[4\times 1,1\times 2,2 \times 3,2\times 4,2\times 3,1\times 2,4\times 1]$, where for each element $N\times S$, $N$ stands for the number of sharing layers and $S$ the share rate for the layer. All experiments were done with a ConvMixer with 768 channels, depth of 32, patch extraction kernel size of 14, and convolutional kernel size of 3.}
    \label{tab:sharing_topologies}
    \resizebox{0.9\columnwidth}{!}{
        \begin{tabular}{cccccc}
        \toprule[1.5pt]
        \multirow{2}{*}{\textbf{Network}} & \textbf{Sharing} & \textbf{Sharing} & \textbf{Params} & \textbf{FLOPs} & \textbf{Top-1} \\
        & \textbf{Distribution} & \textbf{Mapping} & \textbf{(M)} & \textbf{(G)} & \textbf{Acc (\%)} \\
        \midrule[1pt]
        ConvMixer & - & - & 20.46 & 5.03 & 75.71  \\
        \midrule
        \multirow{3}{*}{WS-ConvMixer} & \multirow{3}{*}{Uniform} & Sequential & \multirow{3}{*}{11.02} & \multirow{3}{*}{5.03} & \textbf{73.29} \\
        & & Strided & & & 72.80 \\
        & & Random & & & Diverged \\
        \midrule
        \multirow{2}{*}{WS-ConvMixer} & \multirow{2}{*}{Middle} & Sequential & \multirow{2}{*}{11.02} & \multirow{2}{*}{5.03} & 73.14 \\
        & & Pyramid & & & \textbf{73.22} \\
        \midrule
        \multirow{2}{*}{WS-ConvMixer} & Front & \multirow{2}{*}{Sequential} &  \multirow{2}{*}{11.02} & \multirow{2}{*}{5.03} & \textbf{73.31} \\
        & Back & & & & 72.35 \\
        \bottomrule[1.5pt]
        \end{tabular}
    }
\end{table}
\setlength{\tabcolsep}{1.4pt}

Figure \ref{fig:sharing_topology} visualizes different sharing topologies while Table \ref{tab:sharing_topologies} shows the results of these sharing methods on the ImageNet dataset. When share rate is 2, ConvMixer with uniform-sequential, middle-pyramid, and front-sequential sharing topology result in similar accuracy (2.5\% less than the non-shared model) while other combinations result in lower accuracy. These results are consistent with the layer-wise similarity study in Section \ref{subsec:why_iso}, and suggests that layer-wise similarity may be a reasonable metric for determining which layers to share. Because of the simplicity and flexibility of \emph{uniform-sequential} sharing topology, we use it in the following experiments unless otherwise stated explicitly.
\paragraph{\bfseries Lightweight Dynamic Transformations on Shared Weights.}
\label{subsec:dynamic_weight_transformations}
To improve the performance of a weight shared network, we introduce lightweight dynamic transformations on top of the shared weights for each individual layer. With this, we potentially improve the representational power of the weight sharing network without increasing the parameter count significantly.

To introduce the lightweight dynamic transformation used in this study, we consider a set of $N$ layers to be shared, with a shared weight tensor $W_s$. In the absence of dynamic transforms, the weight tensor $W_s$ would simply be shared among all $N$ layers. We consider $W_{i} \in \mathbb{R}^{C\times C\times K\times K}$ to be the weights of the $i$-th layer, where $C$ is the channel size and $K$ is the kernel size. With a dynamic weight transformation function $f_i$, the weights $W_i$ at the $i$-th layer becomes
\begin{equation}
W_i = f_i(W_s) 
\end{equation}
The choose $f_i$ to be a learnable lightweight affine transformation that allows us to transform the weights without introducing heavy computation and parameter overhead. Specifically, $f_i(W) = \mathbf{a} * W + \mathbf{b}$ applies a grouped point-wise convolution with weights $\mathbf{a} \in \mathbb{R}^{C \times G}$ and bias $\mathbf{b} \in \mathbb{R}^{C}$ to $W$, where $G$ is the number of groups. The number of groups, $G \in [1,C]$, can be varied to modulate the amount of inter-channel mixing.

Table \ref{tab:dynamic_weight_transform} shows the effect of different number of groups in the dynamic weight transformation on the performance and efficiency (in terms of parameters and FLOPs) of ConvMixer on the ImageNet dataset. As Table \ref{tab:dynamic_weight_transform} shows, using $G = 64$, the dynamic weight transformation slightly improves accuracy by 0.07\% (from 73.29\% to 73.36\%) with 7\% more parameters (from 11.02M to 11.8M) and 11.9\% more FLOPs (from 5.03G to 5.63G).
Despite having stronger expressive power, dynamic weight transformation does not provide significant accuracy improvement with under 10\% of overhead on number of weights and FLOPs and sometimes even degrading accuracy.

\newcommand\setrow[1]{\gdef\rowmac{#1}#1\ignorespaces}

\setlength{\tabcolsep}{4pt}
\begin{table}[t!]
    \centering
    \caption{\textbf{Effect of affine transformations on the performance of Weight Shared ConvMixer model with a sharing rate of 2.} All experiments were done with a ConvMixer with 768 channels, depth of 32, patch extraction kernel size of 14, and convolutional kernel size of 3.}
    \label{tab:dynamic_weight_transform}
    \resizebox{0.95\columnwidth}{!}{
        \begin{tabular}{cccccc}
        \toprule[1.5pt]
        \multirow{2}{*}{\textbf{Network}} & \textbf{Weight} & \multirow{2}{*}{\textbf{Group Rate}} & \textbf{Params} & \textbf{FLOPs} & \textbf{Top-1} \\
        & \textbf{Transformation?} & & \textbf{(M)} & \textbf{(G)} & \textbf{Acc (\%)} \\
        \midrule[1pt]
        ConvMixer & - & - & 20.46 & 5.03 & 75.71 \\
        \midrule
        \multirow{5}{*}{WS-ConvMixer} & \xmark & - & 11.02 & 5.03 & 73.29 \\
        & \checkmark & 1 & 11.05 & 5.04 & 72.87 \\
        & \checkmark & 16 & 11.20 & 5.17 & 73.20 \\
        & \checkmark & 32 & 11.40 & 5.31 & 73.14 \\
        & \checkmark & 64 & 11.80 & 5.63 & \textbf{73.36} \\
        \bottomrule[1.5pt]
        \end{tabular}
    }
\end{table}
\setlength{\tabcolsep}{1.4pt}

\paragraph{\bfseries Initializing Weights from Pretrained Non-sharing Networks.}
\label{subsec:fusion_strategies}

Here we consider how we can use the weights of a pretrained, uncompressed network to improve the parameter shared version of an isotropic network. To this end, we introduce transformations on the original weights to generate the weights of the shared network for a given sharing topology. We define $V_j \in \mathbb{R}^{C\times C \times K \times K}$ to be the $j$-th pretrained weight in the original network, and $u_j \in \mathbb{R}^{C}$ to be the corresponding pretrained bias. The chosen sharing topology defines a disjoint set cover of the original network's layers, where each disjoint subset maps a group of layers from the original network to a single shared weight layer. Concretely, if the weight $W_i$ is shared among $S_i$ layers $\{{i_1}, {i_2}, ..., {i_{S_i}}\}$ in the compressed network, then we define $W_i = F_i(V_{i_1}, V_{i_2}, ..., V_{i_{S_i}})$, where we can design each $F_i$. We refer to $F$ as the \textit{fusion strategy}. In all experiments we propagate the gradient back to the original, underlying $V_j$ weights. Importantly, $F$ does not incur a cost at inference-time, since we can constant-fold this function once we finish training.

One simple fusion strategy would be to randomly initialize a single weight tensor for this layer. Note that this is the approach we have used in all previous experiments. We empirically explore the following fusion strategies:

\begin{scriptsize}

\setlength{\tabcolsep}{4pt}
\begin{table}[t!]
    \centering
    \caption{\textbf{Effect of different fusion strategies (Section \ref{subsec:fusion_strategies}) on the performance of ConvMixer.} All experiments were done with a ConvMixer with 768 channels, depth of 32, patch extraction kernel size of 14, and convolutional kernel size of 3. All weight sharing ConvMixer models share groups of 2 sequential layers.}
    \label{tab:pretrained_experiments}
    \resizebox{1.0\columnwidth}{!}{
        \begin{tabular}{ccccc}
        \toprule[1.5pt]
        \multirow{2}{*}{\textbf{Network}}  & \multirow{2}{*}{\textbf{Fusion Strategy}} & \textbf{Params} & \textbf{FLOPs} & \textbf{Top-1} \\
        & & \textbf{(M)} & \textbf{(G)} & \textbf{Acc (\%)} \\
        \midrule[1pt]
        ConvMixer & - & 20.5 & 5.03 & 75.71 \\
        \midrule
        \multirow{6}{*}{WS-ConvMixer} & - & \multirow{6}{*}{10.84} & \multirow{6}{*}{5.03} & 73.23 \\
        & Choose First & & & 74.81 \\
        & Mean & & & 74.91 \\
        & Scalar Weighted Mean & & & \textbf{75.15} \\
        & Channel Weighted Mean & & & \textbf{75.15} \\
        & Pointwise Convoulution & & & Diverged \\
        \bottomrule[1.5pt]
        \end{tabular}
    }
\end{table}
\setlength{\tabcolsep}{1.4pt}

\end{scriptsize}

\begin{itemize}
  \item \textbf{Choose First}: In this setup we take the first of the set of weights within the set: $W_i = F_i(V_{i_1}, V_{i_2}, ..., V_{i_S}) = V_{i_1}$. The choice of the first weight ($V_{i_1}$), rather than any other weight, is arbitrary. Training this method from scratch is equivalent to our vanilla weight sharing strategy.
  \item \textbf{Mean}: We take the average of all the weight tensors within the set, $W_i = \frac{1}{S_i} \sum_{k=1}^{S_i} V_{i_k}$ and $b_i = \frac{1}{S_i} \sum_{k=1}^{S_i} u_{i_k}$.
  \item \textbf{Scalar Weighted Mean}: Same as the average, except each weight tensor gets a learned scalar weighting, $W_i = \frac{1}{S_i} \sum_{k=1}^{S_i} \alpha_{i_k} V_{i_k}$, $\alpha_i \in \mathbb{R}$. We take a simple mean of the bias, just as in the Mean strategy. The idea here is to provide the ability to learn more complex fusions, of which Choose First strategy, and Mean are special cases.
  \item \textbf{Channel Weighted Mean}: Rather than a scalar per layer, each weight tensor has a learned scalar for every filter, $W_i = \frac{1}{S_i} \sum_{k=1}^{S_i} \vec{\alpha_i} V_{i_k}$, $\vec{\alpha_i} \in \mathbb{R}^{C}$. Again, we take a simple mean of the bias, just as the Mean strategy. This strategy should allow the model to choose filters from specific weight tensors, or learn linear combinations.
  \item \textbf{Pointwise Convolution}: In this transformation, a pointwise convolution is applied to each layers weights, that maps to the same size filter, $W_i = \frac{1}{S_i} \sum_{k=1}^{S_i} A_i * V_{i_k}$, $A_i \in \mathbb{R}^{C \times C}$ .This should allow arbitrary mixing and permutations of the kernels of each layer.

\end{itemize}
Table \ref{tab:pretrained_experiments} shows that the Channel Weighted Mean fusion strategy allows us to compress the model by $2 \times$ while maintaining the performance of original network. Furthermore, in Section \ref{sec:representation_analysis}, we show that weight fusion strategies allow us to learn representations similar to the original network.
\section{Effect of Parameter Sharing on Different Isotropic Networks on the ImageNet dataset}
We evaluate the performance of the parameter sharing methods introduced in Section \ref{subsec:design-space-exploration} on a variety of isotropic architectures. For more information on the training set-up and details, see Appendix C. 

\subsection{Parameter Sharing for ConvMixer}
Typically, when considering model scaling, practitioners often vary parameters including the network depth, width, and image resolution, which scale the performance characteristics of the model \cite{efficientnet}. In Table \ref{tab:ws_vs_standard_scaling}, we show that weight sharing models can significantly outperform baselines with the same FLOPs and parameters generated through traditional scaling alone, for example improving accuracy by roughly 10\% Top-1 in some cases. We also show a full family of weight sharing ConvMixer models across multiple architectures in Table \ref{tab:convmixer_ws_family}, and find that weight sharing can reduce parameters by over $2\times$ in many architectures while maintaining similar accuracy. These results  show that weight sharing, in addition to typical scaling methods, is an effective axis for model scaling.

\begin{scriptsize}

\setlength{\tabcolsep}{4pt}
\begin{table}[t!]
    \centering
    \caption{\textbf{Weight sharing vs. model scaling for the ConvMixer model on ImageNet.} For a fair comparison, we generate models with similar FLOPs and network parameters to our family of weight sharing models using traditional model scaling methods. Weight sharing methods achieve significantly better performance than traditional model scaling. See Table \ref{tab:convmixer_ws_family} for more details on the weight sharing model.}

    \label{tab:ws_vs_standard_scaling}
    \resizebox{1.0\columnwidth}{!}{
        \begin{tabular}{ccccccc}
        \toprule[1.5pt]
        \textbf{Network} & \multirow{2}{*}{\textbf{Resolution}} & \textbf{Weight} &\textbf{Share} & \textbf{Params} & \textbf{FLOPs} & \textbf{Top-1} \\
        \textbf{(C/D/P/K)} & & \textbf{Sharing?} &\textbf{Rate} & \textbf{(M)} &\textbf{(G)} & \textbf{Acc(\%)} \\
        \midrule[1pt]
        
        768/32/14/3  & 224 & \xmark & - & 20.5 & 5.03 & 75.71 \\
        \midrule
        576/32/14/3 & 322 & \xmark & - & 11.8 & 5.92 & 70.326 \\
        768/16/14/3 & 322 & \xmark & - & 10.84 & 5.32 & 74.20 \\

        768/32/14/3 & 224 & \checkmark & 2 & 11.02 & 5.03 & \textbf{75.14} \\
        
        \midrule
        384/32/14/3 & 448 & \xmark & - & 5.5 & 5.23 & 58.83 \\
        768/8/14/3 & 448 & \xmark & - & 6.04 & 5.38 & 68.31 \\

        768/32/14/3 & 224 & \checkmark & 4 & 6.3 & 5.03 & \textbf{71.91} \\
        
        \midrule
        288/32/14/3 & 644 & \xmark & - & 3.25 & 6.23 & 40.46 \\
        768/4/14/3 & 644 & \xmark & - & 3.63 & 6.04 & 57.75 \\

        768/32/14/3 & 224 & \checkmark & 8 & 3.95 & 5.03 & \textbf{67.19} \\

        \bottomrule[1.5pt]
        \end{tabular}
    }
\end{table}

\end{scriptsize}
\begin{scriptsize}

\setlength{\tabcolsep}{4pt}
\begin{table}[t!]
    \centering
    \caption{\textbf{Weight sharing family of ConvMixer model on ImageNet.} Significant compression rates can be achieved without loss in accuracy across multiple isotropic ConvMixer models. We also generate a full family of weight sharing models by varying the \textit{share rate}, which is the reduction factor in number of unique layers for the weight shared model compared to the original. C/D/P/K represents the dimension of channel, depth, patch and kernel of the model. If \textit{weight fusion} is specified, the channel weighted mean strategy described in Section \ref{subsec:design-space-exploration} is used.} 
    
    \label{tab:convmixer_ws_family}
    \resizebox{1.0\columnwidth}{!}{
        \begin{tabular}{ccccccc}
        \toprule[1.5pt]
        \textbf{Network} & \textbf{Weight} & \textbf{Share} &  \textbf{Weight} & \textbf{Params} & \textbf{FLOPs} & \textbf{Top-1} \\
        \textbf{(C/D/P/K)} & \textbf{Sharing?} & \textbf{Rate} & \textbf{Fusion?} & \textbf{(M)} &\textbf{(G)} & \textbf{Acc(\%)} \\
        \midrule[1pt]

        \multirow{4}{*}{1536/20/7/3} & \xmark & - & - &  49.4 & \multirow{4}{*}{48.96} & 78.03 \\

        & \checkmark & 2 &  \checkmark & 25.8 & & \textbf{78.47} \\
        & \checkmark & 4 & \checkmark & 14 & & 75.76 \\
        & \checkmark & 10 & \xmark & 6.9 & & 72.27  \\
        
        \midrule
        
        \multirow{4}{*}{768/32/14/3} & \xmark & - & - & 20.5 & \multirow{4}{*}{5.03} & 75.71 \\
        & \checkmark & 2 &\checkmark & 11.02 & & \textbf{75.14} \\
        & \checkmark & 4 &\checkmark & 6.3 & & 71.91 \\
        & \checkmark & 8 &\checkmark & 3.95 & & 67.19 \\
        
        \midrule
        
        \multirow{4}{*}{512/16/14/9} & \xmark & - & - & 5.7 & \multirow{4}{*}{1.33} & 67.48 \\
        & \checkmark & 2  & \checkmark & 3.63 & & \textbf{65.04} \\
        & \checkmark & 4 & \checkmark & 2.58 & & 59.34 \\
        & \checkmark & 8 & \xmark & 2.05 & & 54.25 \\
        
        \bottomrule[1.5pt]
        \end{tabular}
    }
\end{table}

\end{scriptsize}

\subsection{Parameter Sharing for Other Isotropic Networks}
\label{eval:other_iso_networks}
Although our evaluations have focused on ConvMixer, the methods discussed in Section \ref{parameter_sharing_methods} are generic and can be applied to any isotropic model. Here, we show results of applying parameter sharing to ConvNeXt \cite{convnext} and the Vision Transformer (ViT) architecture.

\paragraph{\bfseries ConvNeXt.} Table \ref{tab:convnext_imagenet} shows the results of parameter sharing on the ConvNeXt isotropic architecture. With parameter sharing, we are able to compress the model by $2 \times$  while maintaining similar accuracy on the ImageNet dataset.

\begin{scriptsize}

\setlength{\tabcolsep}{4pt}
\begin{table}[t!]
    \centering
    \caption{\textbf{Effect of weight sharing on the ConvNeXt model on ImageNet.} WS-ConvNeXxt has 2x less number of parameters but still achieves similar accuracy to the original ConvNeXt model.}
    \label{tab:convnext_imagenet}
    
    \resizebox{0.8\columnwidth}{!}{
        \begin{tabular}{cccccc}
        \toprule[1.5pt]
        \multirow{2}{*}{\textbf{Network}} & \multirow{2}{*}{\textbf{Depth}} & \textbf{Share} & \textbf{Params} & \textbf{FLOPs} & \textbf{Top-1} \\
        & & \textbf{Rate} & \textbf{(M)} &\textbf{(G)} & \textbf{Acc(\%)} \\
        \midrule[1pt]
        
        \multirow{2}{*}{ConvNeXt} & 18 & - & 22.3 & 4.3 & 78.7 \\
        & 9 & - & 11.5 & 2.2 & 75.3 \\

        \midrule
        
        \multirow{4}{*}{WS-ConvNeXt} & \multirow{4}{*}{18} & 2 & 11.5 & \multirow{4}{*}{4.3} & \textbf{78.07} \\
        & & 4 & 6.7 &  & 76.11 \\
        & & 6 & 4.3 &  & 72.07 \\
        & & 9 & 3.1 &  & 68.75 \\
        
        \bottomrule[1.5pt]
        \end{tabular}
    }
\end{table}

\end{scriptsize}

\paragraph{\bfseries Vision Transformer (ViT).} 
We also apply our weight sharing method to a Vision Transformer, a self-attention based isotropic network. Due to space limit, we report accuracy numbers in Appendix B. Furthermore, we discuss the differences between applying weight sharing methods to CNNs versus transformers.

\begin{scriptsize}

\setlength{\tabcolsep}{4pt}
\begin{table}[t!]
    \centering
    \caption{{Share rate and ImageNet accuracy comparison with existing weight sharing methods.}}
    \label{tab:vs_existing_weight_sharing}
    \resizebox{0.9\columnwidth}{!}{
        \begin{tabular}{lcccc}
        \toprule[1.5pt]
        \multirow{2}{*}{\textbf{Network}} & \textbf{Share} & \textbf{Params} & \textbf{FLOPs} & \textbf{Top-1} \\
        & \textbf{Rate} & \textbf{(M)} &\textbf{(G)} & \textbf{Acc(\%)} \\
        \midrule[1pt]
        
        ConvMixer-768/32 \cite{convmixer} (baseline) & - & 20.5 & 5.03 & 75.71 \\
        WS-ConvMixer-768/32-S2 (ours) & \textbf{1.86} & \textbf{11.02} & 5.03 & \textbf{75.14} \\
        \midrule
        
        ConvMixer-1536/20 \cite{convmixer} (baseline) & - & 49.4 & 48.96 & 78.03 \\
        WS-ConvMixer-1536/20-S2 (ours) & \textbf{1.91} & \textbf{25.8} & 48.96 & \textbf{78.47} \\
        \midrule
        
        ConvNeXt-18 \cite{convnext} (baseline) & - & 22.3 & 4.3 & 78.7 \\
        WS-ConvNeXt-18-S2 (ours) & \textbf{1.92} & \textbf{11.5} & 4.3 & \textbf{78.07} \\
        WS-ConvNeXt-18-S4 (ours) & 3.33 & 6.7 & 4.3 & 76.11 \\
        \midrule
        
        ResNet-152 \cite{resnet} (baseline) & - & 60 & 11.5 & 78.3 \\
        IamNN \cite{iamnn} & 12 & 5 & 2.5-9 & 69.5 \\
        \midrule
        
        ResNet-101 \cite{resnet} (baseline) & - & 44.54 & 7.6 & 77.95 \\
        DR-ResNet-65 \cite{Guo2019DynamicRN} & 1.58 & 28.12 & 5.49 & 78.12 \\ 
        DR-ResNet-44 \cite{Guo2019DynamicRN} & 2.2 & 20.21 & 4.25 & 77.27 \\
        \midrule
        
        ResNet-50 \cite{resnet} (baseline) & - & 25.56 & 3.8 & 76.45 \\
        DR-ResNet-35 \cite{Guo2019DynamicRN} & 1.45 & 17.61 & 3.12 & 76.48 \\ 
        ResNet50-OrthoReg \cite{deeplyshared} & 1.25 & 20.51 & 4.11 & 76.36 \\
        ResNet50-OrthoReg-SharedAll \cite{deeplyshared} & 1.6 & 16.02 & 4.11 & 75.65 \\
        
        \bottomrule[1.5pt]
        \end{tabular}
    }
\end{table}

\end{scriptsize}
\subsection{Comparison with  State-of-the-art Weight Sharing Methods.}
Table \ref{tab:vs_existing_weight_sharing} compares the performance of weight sharing methods discussed in Section \ref{subsec:fusion_strategies} with existing methods \cite{deeplyshared, Guo2019DynamicRN, iamnn} on ImageNet. Compared to existing methods, our weight sharing schemes are effective; achieving higher compression rate while maintaining accuracy. For example, ConvMixer-768/32, ConvMixer-156/20, and ConvNeXt-18 with weight sharing and weight fusion achieve 1.86x, 1.91x and 1.92 share rate while having a similar accuracy. Existing weight sharing techniques \cite{deeplyshared, Guo2019DynamicRN} can only achieve at most 1.58x and 1.45x share rate at while maintaining accuracy.
Although \cite{iamnn} can achieve 12x share rate, it results in a 8.8\% accuracy drop.

These results show that isotropic networks can achieve a high share rate while maintaining accuracy with simple weight sharing methods. The traditional pyramid style networks, while using complicated sharing schemes \cite{deeplyshared, Guo2019DynamicRN, iamnn}, the share rate is usually limited. Note that although our sharing schemes can achieve higher share rates, existing methods like \cite{iamnn, deeplyshared} are able to directly reduce FLOPs, which our method does not address.
\section{Representation Analysis} \label{sec:representation_analysis}
\begin{figure}[th!]
    \centering
    \begin{subfigure}{0.48\linewidth}
        \includegraphics[width=\linewidth]{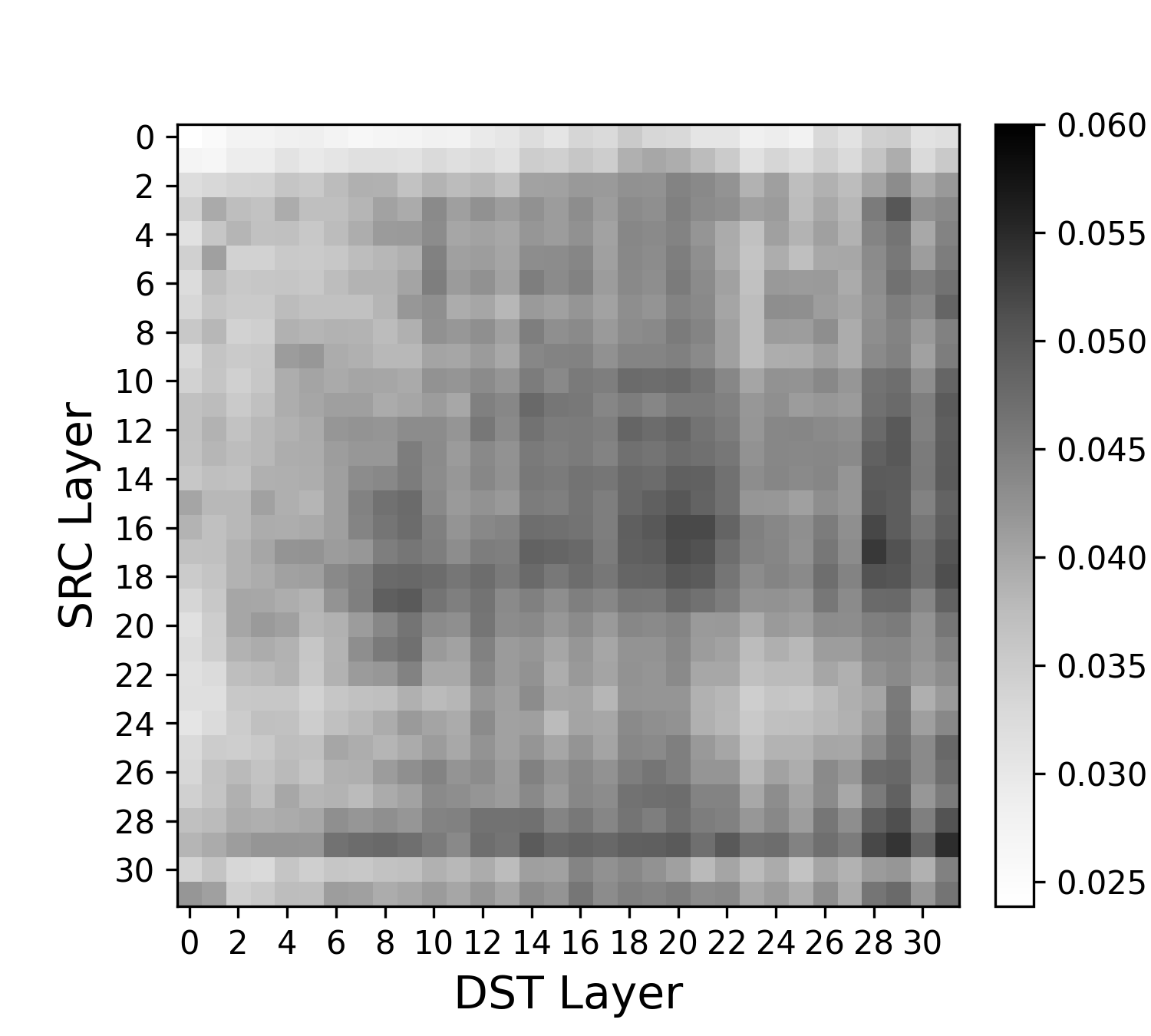}
        \caption{Vanilla WS-ConvMixer.}
        \label{subfig:}
    \end{subfigure}
    \begin{subfigure}{0.48\linewidth}
        \includegraphics[width=\linewidth]{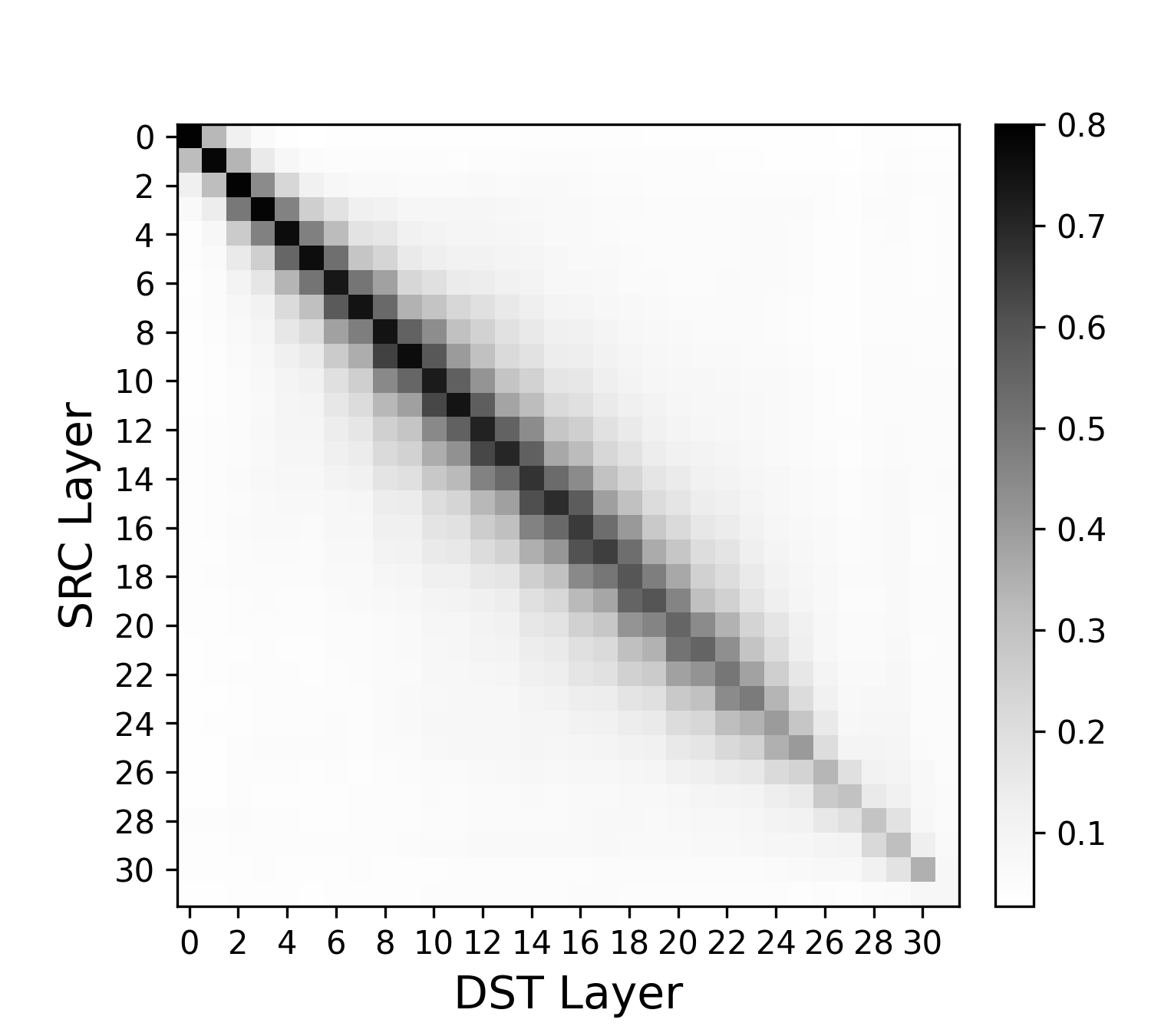}
        \caption{WS-ConvMixer with Fusion.}
        \label{subfig:}
    \end{subfigure}
    \caption{(a) The CKA similarity analysis of a standard ConvMixer's intermediate feature maps compared to a vanilla weight shared ConvMixer, with share rate of 2. (b)
    The same analysis but compare to a weight shared ConvMixer initialized with weight fusion. The channel weighted mean fusion strategy is used (see Section \ref{subsec:design-space-exploration}).}
    \label{fig:wsconvmixer_vs_convmixer_cka}
\end{figure}

In this sections, we perform qualitative analysis of our weight sharing models to better understand why they lead to improved performance and how they change model behavior. To do this, we first analyze the representations learned by the original network, compared to one trained with weight sharing. We follow a similar set-up to Section \ref{subsec:why_iso}. We use CKA as a metric for representational similarity and compute pairwise similarity across all layers in both the networks we aim to compare. In Figure \ref{fig:wsconvmixer_vs_convmixer_cka}(a) we first compare the representations learned by a vanilla weight sharing method to the representations of the original network. We find that there is no clear relationship between the representations learned. Once we introduce the weight fusion initialization strategy (Section \ref{subsec:design-space-exploration}), we find significant similarity in representations learned, as shown in Figure \ref{fig:wsconvmixer_vs_convmixer_cka}(b). This suggests that our weight fusion initialization can guide the weight shared models to learn similar features to the original network. In Appendix F, we further analyze the weight shared models and characterize their robustness compared to standard networks.

\section{Conclusion}
Isotropic networks have the unique property in which all layers in the model have the same structure, which naturally enables parameter sharing. In this paper, we perform a comprehensive design space exploration of shared parameters in isotropic networks (SPIN), including the weight sharing topology, dynamic transformations and weight fusion strategies. Our experiments show that, when applying these techniques, we can compress state-of-the-art isotropic networks by up to 2 times without losing any accuracy across many isotropic architectures. Finally, we analyze the representations learned by weight shared networks and qualitatively show that the techniques we introduced, specifically fusion strategies, guide the weight shared model to learn similar representations to the original network. These results suggest that parameters sharing is an effective axis to consider when designing efficient isotropic neural networks.

\bibliographystyle{splncs04}
\bibliography{egbib}

\newpage
\appendix
\section{Weight Sharing Search Space Characterization}\label{appendix:searchspace}

\subsection{Isotropic Network Case}

Suppose we have an $L$ layer isotropic network and a weight tensor budget of $P \in \mathbb{Z}$, where $0 < P \leq L$ (recall that $\frac{L}{P}$ is the \textit{share rate}).
When building our network, we can choose between $P$ different weight tensors for each layer (sampling with replacement), so there are $P^L$ choices. Thus the search space for parameter sharing can be described as $\Omega(L, P) = P^L$.

We can define the size of the search space of topologies which use \textit{exactly} $P$ parameter tensors as $\Tilde{\Omega}(L, P) = \Omega(L, P) - \Omega(L, P - 1)$. This simply reduces the size of the original search space by the number of topologies which have up to $P - 1$ shared weight tensors.

\subsection{``Staged" Network Case}

Suppose we have a network with $L_N$ total layers, but with $S$ discrete stages (or, more generally, disjoint subsets of layers), where the weight tensors have identical shape only to other weight tensors in their respective stage (or subset). This is a common paradigm for many popular CNN backbone architectures, such as ResNet \cite{resnet}, MobileNet \cite{Howard2017MobileNetsEC}, and DenseNet \cite{huang2018densely}. Without loss of generality, we refer to all disjoint subset architectures as staged architectures.

We simplify the following analysis by assuming each stage of the network has exactly $L_S$ layers. Then we define $\Omega_S(L_N, L_S, P)$ to be the number of non-degenerate topologies for a staged network, staying below the $P$ weight tensor budget:
\begin{equation}
\Omega_S(L_N, L_S, P) = \left\{
\begin{array}{ll}
    0, L_N \leq 0 \vee P \leq 0 \\
    \sum_{i=1}^{\text{min}(L_S,P)} \binom{P}{i} \Tilde{\Omega}(L_S, i) \Omega_S(L_N - L_S, L_S, P - i) \\
\end{array} 
\right.
\end{equation}

\begin{figure}[th!]
    \centering
    \includegraphics[width=0.5\linewidth]{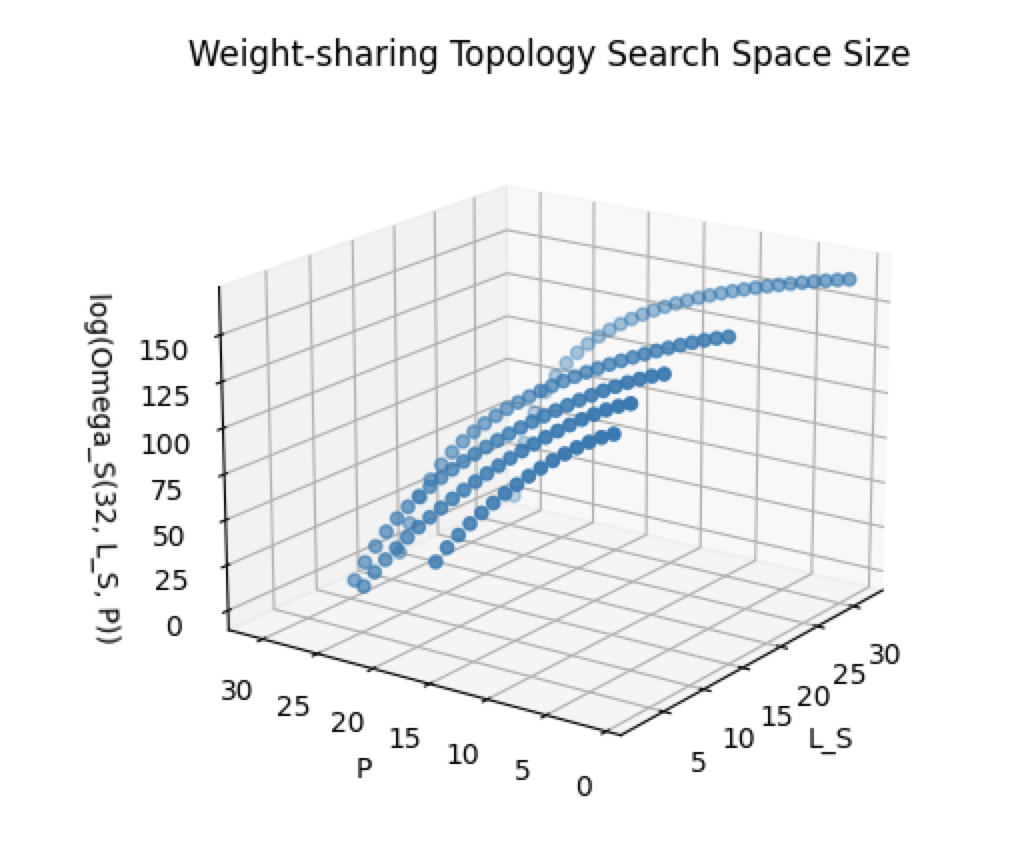}
    \caption{Log plot of search space sizes for various $L_S$ and $P$ values for a depth $L_N=32$ network. Recall that larger $L_S$ values correspond to a ``more isotropic'' architecture.}
    \label{fig:searchspace}
\end{figure}

It can be shown numerically that $\Omega_S(L_N, L_S, P)$ increases rapidly as a function of $L_S$ (see Figure \ref{fig:searchspace} for a graphical representation). In other words, architectures closer to isotropic architectures support more options for parameter sharing, when the overall number of layers is held constant.

\section{Weight Sharing in Vision Transformers (ViTs)}
\label{appendix:ws-vits}
Table \ref{tab:ws_vits} shows results of parameter sharing for the DeiT architecture, which is a popular variant of ViT.  
We experiment with plain weight sharing and weight fusion techniques described in Section 3.2 on the DeiT-Ti and DeiT-S model, both of which have 12 layers. 

Compared to the baseline, the WS-DeiT-S model is able to get similar accuracy, within 1 point Top-1 with half the parameters. At iso-parameters, the WS-DeiT models significantly outperform non-weight-sharing models. Among the weight sharing schemes, using weight fusion with pretrained weights can consistently boost accuracy. For example, using weight fusion can increase accuracy 0.55\% on WS-Deit-Ti and 0.83\% on WS-Deit-S compared to the plain weight sharing version.

\begin{scriptsize}
\begin{table}[t!]
    \caption{\textbf{DeiT Top-1 accuracy on ImageNet-1k with different weight fusion and share rates.} All trained models are based off of DeiT-Ti (which has 12 transformer layers). WS-DeiT stands for the weights sharing version. All experiments use the \textbf{Sequential} sharing topology. The method with the clearly highest performance is bold-faced for each parameter regime.}
    \vspace{5pt}
    \label{tab:vit}
        \begin{tabular}{cccccccc}
        \toprule[1.5pt]
        
        \multirow{2}{*}{\textbf{Model}} & \multirow{2}{*}{\textbf{Depth}} & \textbf{Fusion} & \textbf{Share} & \textbf{Params} & \textbf{FLOPs} & \textbf{ImgNet} \\
        & & \textbf{Strategy} & \textbf{Rate} & \textbf{(M)} & \textbf{(G)} & \textbf{Acc(\%)} \\
        
        \midrule[1pt] 
        
        \multirow{3}{*}{Deit-Ti} & 12 & - & - & 5.72 & 1.26 & 72.55 \\
        & 6 & - & - & 3.05 & 0.64 & 63.11 \\
        & 4 & - & - & 2.16 & 0.44 & 55.61 \\
        
        \midrule 
        
        \multirow{2}{*}{WS-DeiT-Ti} & \multirow{2}{*}{12} & \multirow{2}{*}{-} & 2 & 3.05 & 1.26 & 68.07 \\
        & & & 3 & 2.16 & 1.26 & 63.50 \\
        
        \midrule 
        
        \multirow{2}{*}{WS-DeiT-Ti} & \multirow{2}{*}{12} & \multirow{2}{*}{Mean} & 2 & 3.05 & 1.26 & 67.96 \\
        & & & 3 & 2.16 & 1.26 & \textbf{63.75} \\
        
        \midrule 
        
        \multirow{2}{*}{WS-DeiT-Ti} & \multirow{2}{*}{12} & \multirow{2}{*}{Scalar Weighted Mean} & 2 & 3.05 & 1.26 & \textbf{68.62} \\
        & & & 3 & 2.16 & 1.26 & 63.74 \\
        
        \midrule[1pt] 
        
        \multirow{3}{*}{Deit-S} & 12 & - & - & 22.05 & 4.61 & 80.52 \\
        & 6 & - & - & 11.40 & 2.33 & 74.48 \\
        & 4 & - & - & 7.85 & 1.58 & 67.77 \\
        
        \midrule 
        
        \multirow{2}{*}{WS-DeiT-S} & \multirow{2}{*}{12} & \multirow{2}{*}{-} & 2 & 11.41 & 4.61 & 78.61 \\
        & & & 3 & 7.87 & 4.61 & 76.67 \\
        
        \midrule 
        
        \multirow{2}{*}{WS-DeiT-S} & \multirow{2}{*}{12} & \multirow{2}{*}{Mean} & 2 & 11.41 & 4.61 & \textbf{79.44} \\
        & & & 3 & 7.87 & 4.61 & \textbf{77.11} \\
        
        \bottomrule[1.5pt] 
        \end{tabular}
    \label{tab:ws_vits}
\end{table}
\end{scriptsize}

\section{Training Details} 
\label{appendix:training_details}
We describe the training details we used to produce the experiments throughout this paper. To produce baseline accuracy of ConvMixer \cite{convmixer}, DeiT \cite{deit} and ConvNeXt \cite{convnext}, we follow the default training settings described in each model's original paper and the released source code as closely as possible.

We train all models on ImageNet-1K dataset without additional data. For ConvMixer, we use a learning rate of 0.01 and batch size of 64 for each GPU. The data augmentations we use includes RandAugment, MixUp, CutMix and random erasing. We use the AdamW optimizer with weight decay 2e-5 and a cosine learning rate schedule with a single cycle. For ConvMixer-1536/20/7/3, we train 150 epochs. For ConvMixer-768/32/14/3 and 512/16/14/9, we train for 300 epochs. For the Weight Sharing ConvMixer models, we use exact the same training setting as each corresponding baseline model to train. It is worth noting that we are able to show that the weight sharing ConvMixer can perform well without any parameter tuning, but this architecture may have a different optimal setting for these hyper parameters, for example due to having less parameters, and proper tuning may further boost performance of our method.

For DeiT and ConvNeXt, we follow the same settings proposed in the respective papers. All DeiT variants were trained with an effective batch size of 256 on 4 GPUs (note that the learning rate is scaled appropriately according to their scaling rule).

\section{Ablation on Sharing Different Components in ConvMixer Block}
For the ConvMixer architecture, there are many components in each block we can choose whether to share. These components include Pointwise and Depthwise Convolution layer, bias for each Convolution layer, and the BatchNorm layer.
In Table \ref{tab:compare_pwise}, we provide a full ablation study on sharing all the components, and gradually turn-off sharing on BatchNorm, Bias, and Depthwise Convolution, in order of the number of parameters each component contains.

As Table \ref{tab:compare_pwise} shows, Pointwise Convolution layer contains the majority of the parameters of each block and only sharing Pointwise Convolution layer results in the best accuracy. Therefore, in our main study, we only share weights for Pointwise Convolution layers for ConvMixer models.

\setlength{\tabcolsep}{4pt}
\begin{scriptsize}
\begin{table}[t!]
    \centering
    \caption{\textbf{Top-1 Accuracy on ImageNet ablating which operation within a ConvMixer Block are shared.} We consider sharing the BatchNorm, Bias of convolutional layer, Depthwise Convolution, and Pointwise Convolution. The model size used in this comparison is 768/32/14/3 for channel/depth/patch size/kernel size. The sharing rate is 2. We apply no transformation or weight fusion in this study.}
    \label{tab:compare_pwise}
    \resizebox{0.9\columnwidth}{!}{
        \begin{tabular}{cccccccc}
        \toprule[1.5pt]
        \multirow{2}{*}{Network} & Share & Share & Share & Share & Params & FLOPs & ImgNet \\
        & BN & Bias & Dwise & Pwise & (M) & (G) & Acc(\%) \\
        \midrule[1pt]
        ConvMixer & \xmark & \xmark & \xmark & \xmark & 20.5 & 5.03 & 74.93 \\
        \midrule
        \multirow{4}{*}{WS-ConvMixer} & \checkmark & \checkmark & \checkmark & \checkmark & 10.84 & \multirow{4}{*}{5.03} & Diverged \\
        & \xmark & \checkmark & \checkmark & \checkmark & 10.89 & & 73.17 \\
        & \xmark & \xmark & \checkmark & \checkmark & 10.92 & & 73.19 \\
        & \xmark & \xmark & \xmark & \checkmark & 11.02 & & 73.29 \\
        \bottomrule[1.5pt]
        \end{tabular}
    }
\end{table}
\end{scriptsize}

\section{Ablation on Weight Fusion Networks without Utilizing Pretrained Weights.}
In Section 3.2, we described weight fusion methods to fuse a pretrained network's weights as initialization for a weight sharing model and showed accuracy improvement. Such weight fusion methods can also be applied without using a pretrained network's weights. To further understand the effect of the proposed weight fusion techniques, we perform an ablation study on applying the same weight fusion strategies with networks that are regularly initialized. 

As results of the ablation study in Table \ref{tab:ablation_pretrained_weights} show, applying weight fusion to a randomly initialized model does not improve accuracy. It is worth noting that, after fusion, the number of effective weights will be the same as a vanilla weight sharing model, so there is no increase in representational power. This ablation study empirically shows that using the fused weights from a pretrained network to initialize a weight sharing model (see Section 3.2) is what leads to improved accuracy, rather than the weight sharing fusion alone.

\begin{scriptsize}

\setlength{\tabcolsep}{4pt}
\begin{table}[t!]
    \centering
    \caption{\textbf{Ablations on whether using a pretrained network to initialize a weight sharing network when using weight fusion.} All experiments were done with a ConvMixer with 768 channels, depth of 32, patch extraction kernel size of 14, and convolutional kernel size of 3. All weight sharing ConvMixer models share groups of 2 sequential layers. We used slightly different hyperparameters for this ablation study and have slightly higher accuracy for WS-ConvMixers.}
    \label{tab:ablation_pretrained_weights}
    \resizebox{1.0\columnwidth}{!}{
        \begin{tabular}{cccccc}
        \toprule[1.5pt]
        \multirow{2}{*}{\textbf{Network}} & \textbf{Weight} & \textbf{Weight} & \textbf{Params} & \textbf{FLOPs} & \textbf{Top-1} \\
        & \textbf{Init.} & \textbf{Fusion} & \textbf{(M)} & \textbf{(G)} & \textbf{Acc (\%)} \\
        \midrule[1pt]
        ConvMixer & Regular & \xmark & 20.5 & 5.03 & 75.71 \\
        \midrule
        \multirow{3}{*}{WS-ConvMixer} & Regular & \xmark & \multirow{3}{*}{10.84} & \multirow{3}{*}{5.03} & 74.29 \\
        & Regular & Choose First & & & 74.25 \\
        & Regular & Mean & & & 74.25 \\
        \midrule
        \multirow{2}{*}{WS-ConvMixer} & Pretrained & Choose First & \multirow{2}{*}{10.84} & \multirow{2}{*}{5.03} & 74.88 \\
        & Pretrained & Mean & & & 75.28 \\
        \bottomrule[1.5pt]
        \end{tabular}
    }
\end{table}
\end{scriptsize}

\section{Further Model Analysis} \label{appendix:model_generalization}
In this section, we provide analysis to further understand why weight sharing is effective for isotropic architectures. All analysis below is performed on the ConvMixer architecture. We first analyze the robustness of these models, followed by further representation analysis with Centered Kernel Alignment (CKA).

\subsection{Robustness to Label Noise}
\begin{figure}[th!]
    \centering
        \includegraphics[width=\linewidth]{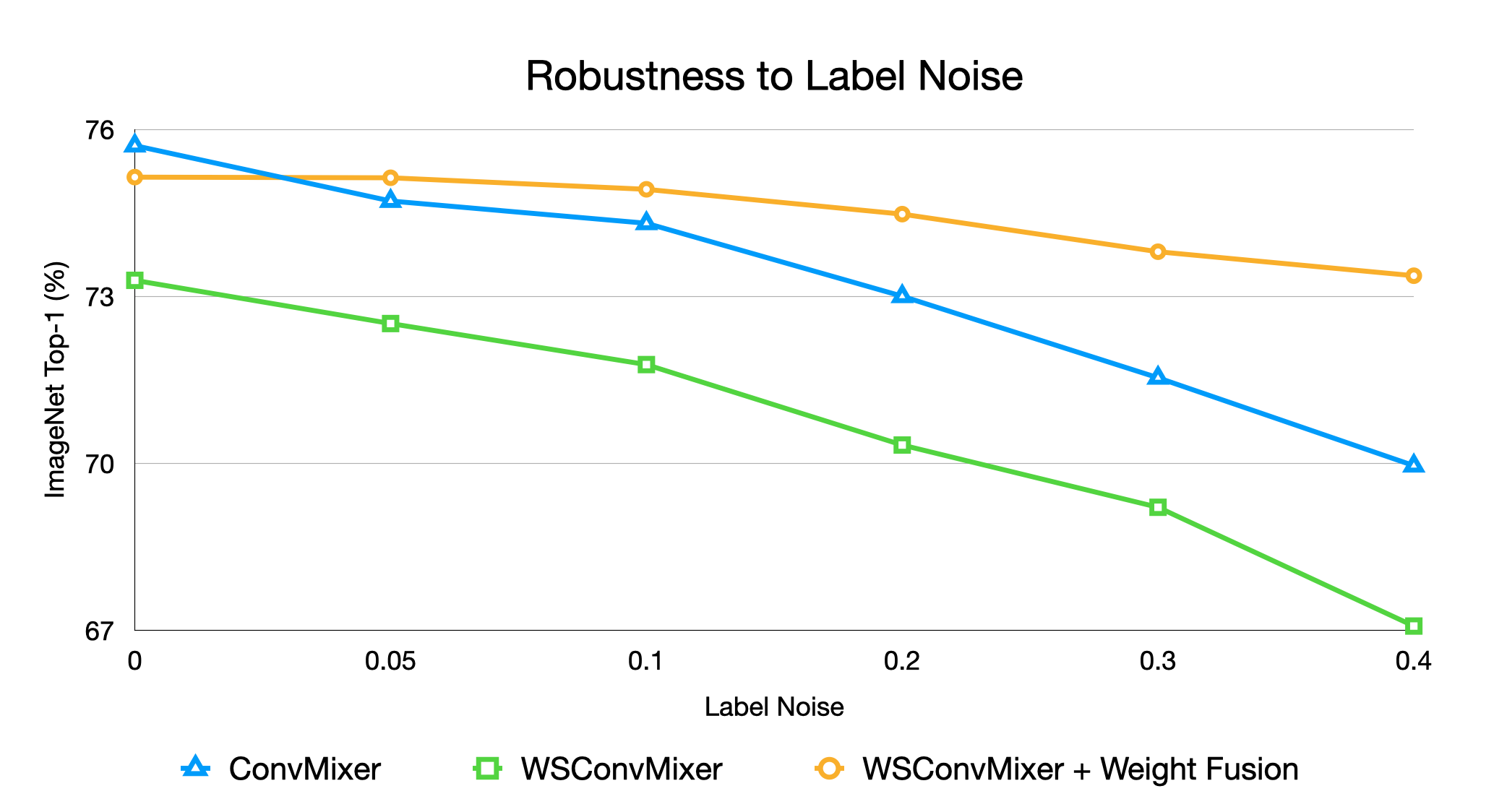}
        \caption{
        Label noise analysis of ConvMixers. 
        The model size used in this comparison is 768/32/14/3 for channel/depth/patch size/kernel size. At any label noise level above zero, the WSConvMixer + WeightFusion model outperforms the baseline ConvMixer, with half the parameters and iso FLOP. This suggests that our weight sharing method generates models that are more robust to noise, and this may be part of the reason we find empirically compelling results.
        }
    \label{fig:label_noise}
\end{figure}

We analyze the robustness of our weight sharing model in the presence of label noise. Our analysis follows \cite{learning-subspaces}. We first choose a noise level $l \in [0, 1]$. This corresponds to the fraction of training image labels to adjust. We then randomly choose a fraction $l$ of images from the training set and set their label to a random category label. We then proceed with training as usual. Note that the labels are unchanged after their initial alteration at the beginning of training. We do not modify the evaluation set in any way. In Figure \ref{fig:label_noise} we show that our weight sharing ConvMixer using the weight fusion method described in Section 3.2 is significantly more robust to noise than the baseline ConvMixer. We are able to exceed the baseline model in absolute Top-1 at all noise levels above zero, while halving the parameters of the model and iso-FLOPs. These results suggest that our weight sharing models provide increased robustness, and part of our empirical improvements in accuracy may be attributable to this property.

\begin{figure}[th!]
    \centering
    \includegraphics[width=1\linewidth]{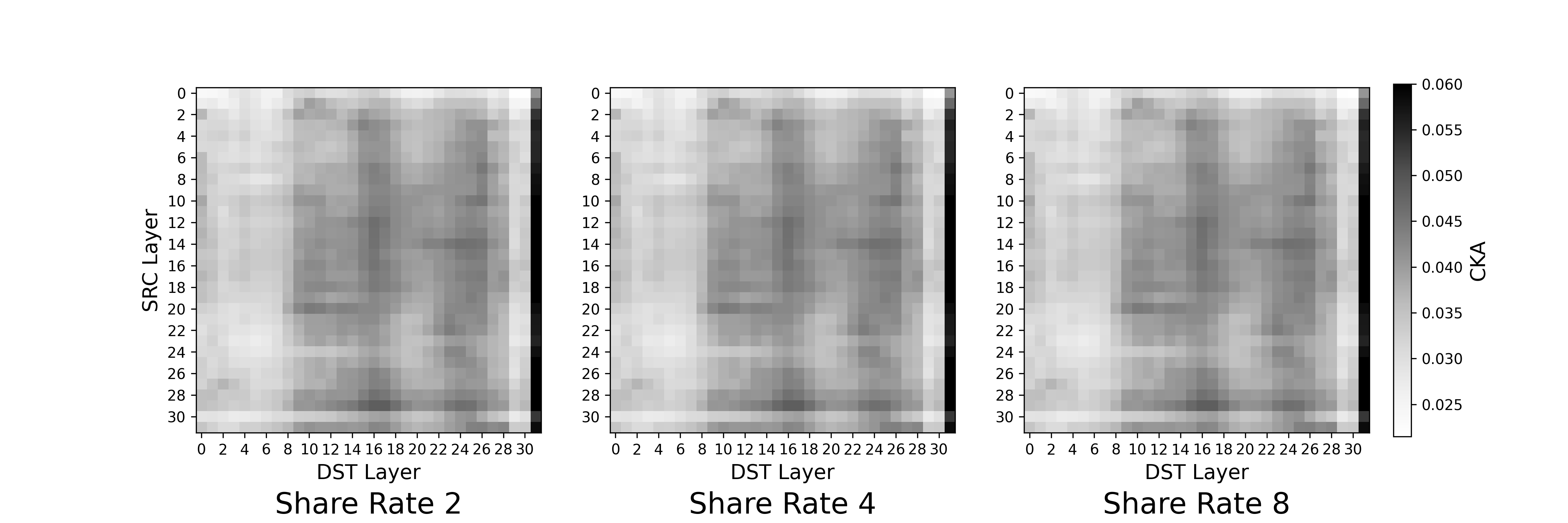}
    \caption{CKA analysis on vanilla WS-ConvMixer-768/32/14/3 model. We compute pairwise analysis of WS-ConvMixer layer feature maps to the non-sharing ConvMixer model using CKA. For this WS-ConvMixer model, it has 73.29\%, 70.11\%, 66.31\% accuracy on ImageNet for share rates 2, 4, and 8 respectively}
    \label{fig:Vanilla-WSConvMixer-CKA}
\end{figure}

\begin{figure}[th!]
    \centering
    \begin{subfigure}{1\linewidth}
        \includegraphics[width=\linewidth]{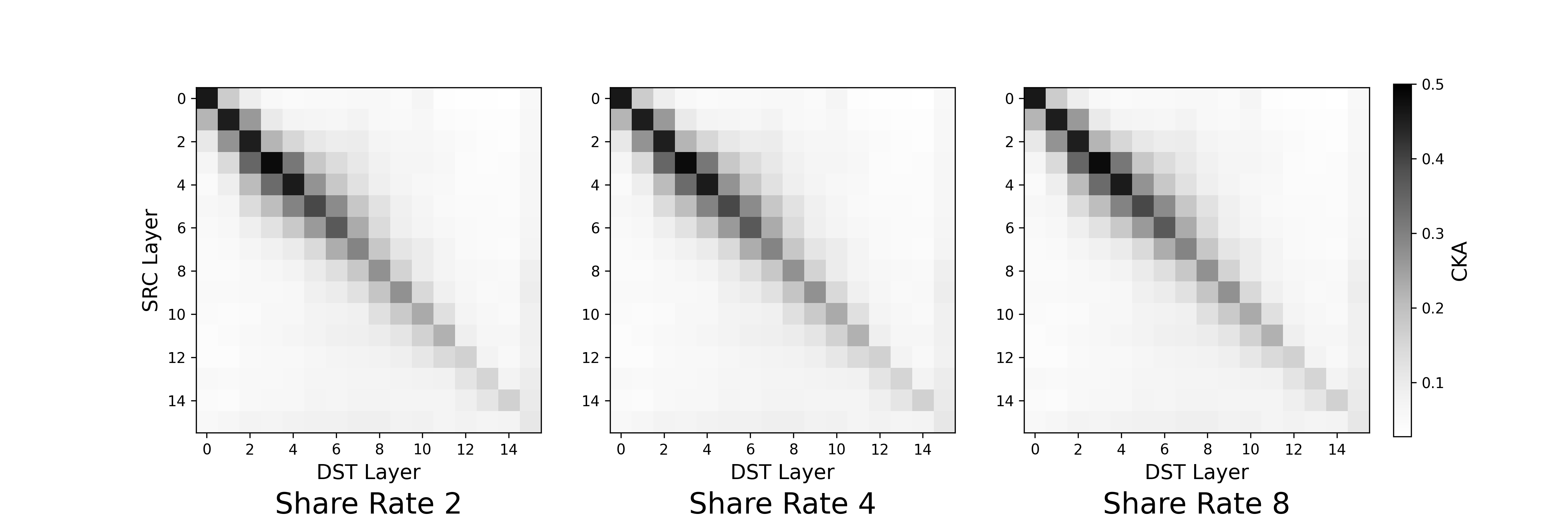}
        \caption{WSConvMixer-512/16 with Weight Fusion. The original ConvMixer has 67.48\% accuracy on ImageNet. For WS-ConvMixer with Weight Fusion, it has 65.04\%, 59.34\%, and 52.95\% accuracy on ImageNet.}
        \label{subfig:}
    \end{subfigure}
    \begin{subfigure}{1\linewidth}
        \includegraphics[width=\linewidth]{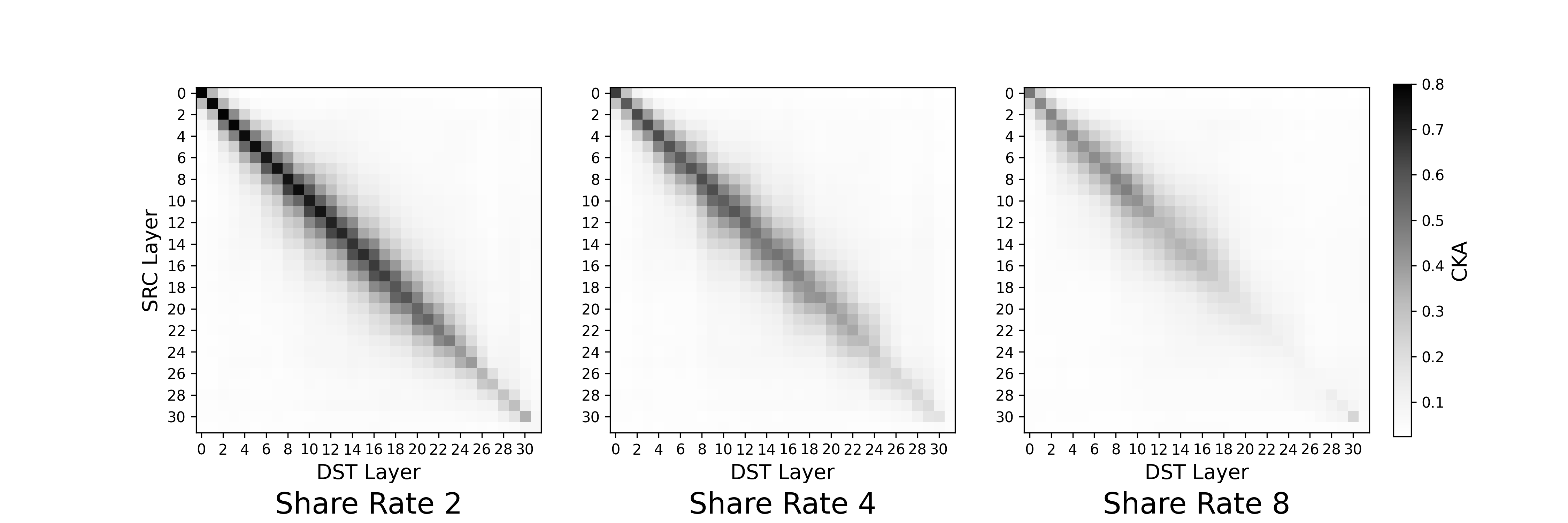}
        \caption{WSConvMixer-768/32 with Weight Fusion. The original ConvMixer has 75.71\% accuracy on ImageNet. For WS-ConvMixer with Weight Fusion, it has 75.14\%, 67.15\%, and 59.69\% accuracy on ImageNet.}
        \label{subfig:}
    \end{subfigure}
    \begin{subfigure}{1\linewidth}
        \includegraphics[width=\linewidth]{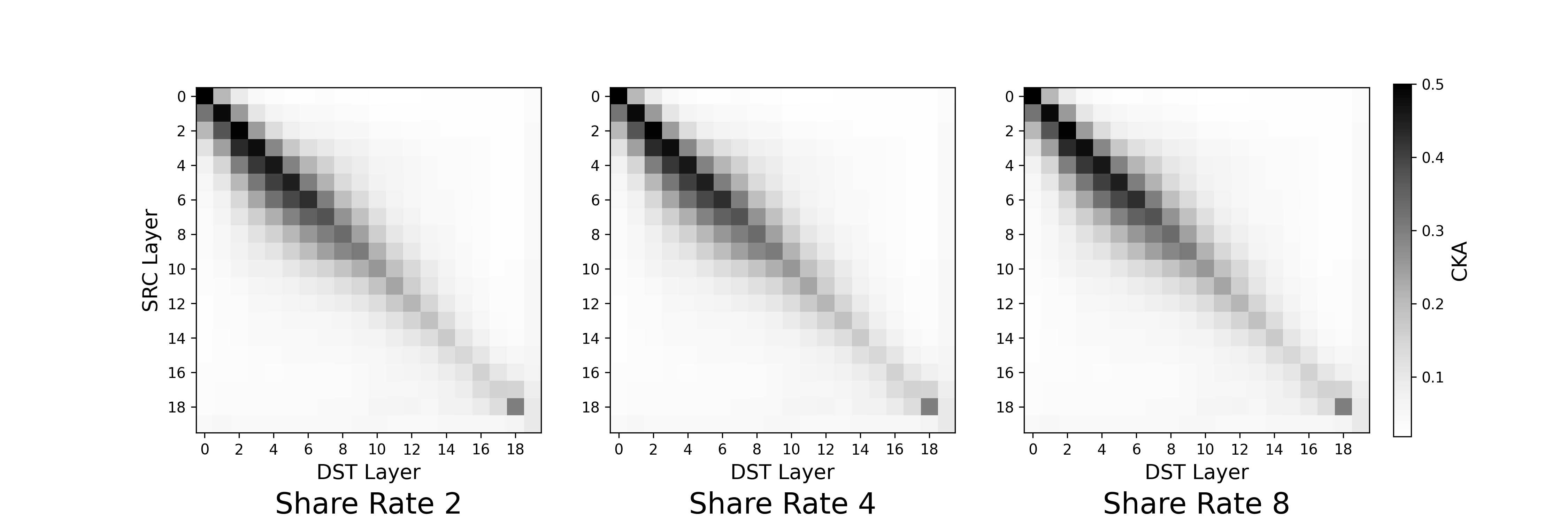}
        \caption{WSConvMixer-1536/20 with Weight Fusion. The original ConvMixer has 78.03\% accuracy on ImageNet. For WS-ConvMixer with Weight Fusion, it has 78.47\%, 75.76\%, and 71.4\% accuracy on ImageNet.}
        \label{subfig:}
    \end{subfigure}
    \caption{CKA analysis on WS-ConvMixer model with Weight Fusion. We compute pairwise analysis of WS-ConvMixer layer feature maps to the non-sharing ConvMxier model using CKA}
    \label{fig:Pretrained-WSConvMixer-CKA}
\end{figure}

\subsection{Further CKA Analysis on WS-ConvMixer}
In this section, we provide more representational analysis using CKA on the Weight Sharing (WS) ConvMixer. In Figure \ref{fig:Vanilla-WSConvMixer-CKA}, we show that for various share rates, the vanilla WS-ConvMixer does not have any clear pattern of layer-wise representational similarity with the original ConvMixer model (768/32/14/3 architecture setting). It's also worth noting the absolute value of similarity is quite low for these models. On the other hand, in Figure \ref{fig:Pretrained-WSConvMixer-CKA}, we show that in the WS-ConvMixer models with weight fusion we see a clear relationship in the representations learned by the weight sharing model compared to the original, as well as significantly higher absolute similarity. This trend holds across multiple architecture settings (2/16/14/9, 768/32/14/3 and 1536/20/14/3) and share rates (2, 4, 8). This finding suggests that weight sharing models have the ability to generate similar representations to the original models, even with significantly less parameters, but need advanced training methods such as our weight fusion strategy to achieve this in practice.

\end{document}